%% file: main.tex
\documentclass{article}

\usepackage{microtype}
\usepackage{graphicx}
\usepackage{subfigure}
\usepackage{booktabs} 
\usepackage{makecell}
\usepackage{amsmath}
\usepackage[usenames, dvipsnames]{color}
\usepackage{graphicx}
\usepackage{wrapfig}
\usepackage{multirow}
\usepackage[inline]{enumitem}
\usepackage{amssymb}
\usepackage{setspace}
\usepackage{makecell}
\usepackage{textcomp}
\usepackage[skip=\baselineskip]{caption}
\usepackage{sidecap}
\usepackage{afterpage}

\usepackage{soul}
\newcommand{\ms}[2]{#1\tiny{$\pm$#2}}

\usepackage{hyperref}

\usepackage[accepted]{icml2019}

\icmltitlerunning{Learning What and Where to Transfer}

\begin{document}

\twocolumn[
\icmltitle{Learning What and Where to Transfer}



\icmlsetsymbol{equal}{*}

\begin{icmlauthorlist}
\icmlauthor{Yunhun Jang}{equal,kaist,omnious}
\icmlauthor{Hankook Lee}{equal,kaist}
\icmlauthor{Sung Ju Hwang}{kaist-cs,kaist-ai,aitrics}
\icmlauthor{Jinwoo Shin}{kaist,kaist-ai,aitrics}
\end{icmlauthorlist}

\icmlaffiliation{kaist}{School of Electrical Engineering, KAIST, Korea}
\icmlaffiliation{kaist-cs}{School of Computing, KAIST, Korea}
\icmlaffiliation{kaist-ai}{Graduate School of AI, KAIST, Korea}
\icmlaffiliation{aitrics}{AITRICS, Korea}
\icmlaffiliation{omnious}{OMNIOUS, Korea}

\icmlcorrespondingauthor{Jinwoo Shin}{jinwoos@kaist.ac.kr}

\icmlkeywords{Machine Learning, ICML}

\vskip 0.3in
]



\printAffiliationsAndNotice{\icmlEqualContribution} 

\begin{abstract}
  As the application of deep learning has expanded to real-world problems with insufficient volume of training data, transfer learning recently has gained much attention as means of improving the performance in such small-data regime. However, when existing methods are applied between heterogeneous architectures and tasks, it becomes more important to manage their detailed configurations and often requires exhaustive tuning on them for the desired performance. To address the issue, we propose a novel transfer learning approach based on meta-learning that can automatically learn what knowledge to transfer from the source network to where in the target network. Given source and target networks, we propose an efficient training scheme to learn meta-networks that decide (a) which pairs of layers between the source and target networks should be matched for knowledge transfer and (b) which features and how much knowledge from each feature should be transferred. We validate our meta-transfer approach against recent transfer learning methods on various datasets and network architectures, on which our automated scheme significantly outperforms the prior baselines that find ``what and where to transfer'' in a hand-crafted manner.
\end{abstract}

\input{Introduction.tex}

\input{Methods.tex}

\input{Experiments.tex}

\input{Conclusion.tex}

\section*{Acknowledgements}

This work was supported by Institute for Information \& communications
Technology Promotion (IITP) grant funded by the Korea government MSIT
(No.2016-0-00563, Research on Adaptive Machine Learning Technology
Development for Intelligent Autonomous Digital Companion) and
supported by the Engineering Research Center Program through the National Research Foundation of Korea (NRF)
funded by the Korean Government MSIT (NRF-2018R1A5A1059921).

\bibliography{References}
\bibliographystyle{icml2019}

\input{supp.tex}





\end{document}

%% file: Introduction.tex
\section{Introduction}
Learning deep neural networks (DNNs) requires large datasets, but it is expensive to collect a sufficient amount of labeled samples for each target task. A popular approach for handling such lack of data is transfer learning \cite{pan2010survey}, whose goal is to transfer knowledge from a known source task to a new target task.
The most widely used method for transfer learning is {\it pre-training with fine-tuning} \cite{razavian2014cnn}:
first train a source DNN (e.g. ResNet \cite{he2016deep}) with a large dataset
(e.g. ImageNet \cite{deng2009imagenet}) and then, use the learned weights
as an initialization to train
a target DNN.
Yet, fine-tuning definitely is not a panacea. If the source and target tasks are semantically distant,
it may provide no benefit. 
Cui et al.~\yrcite{cui2018large} suggest to sample from the source dataset depending on a target task
for pre-training,
but it is only possible when the source dataset is available.
There is also no straightforward way to use fine-tuning, if the network architectures
for the source and target tasks largely differ.

\begin{figure}[t]
\begin{center}
\centerline{\includegraphics[width=\columnwidth]{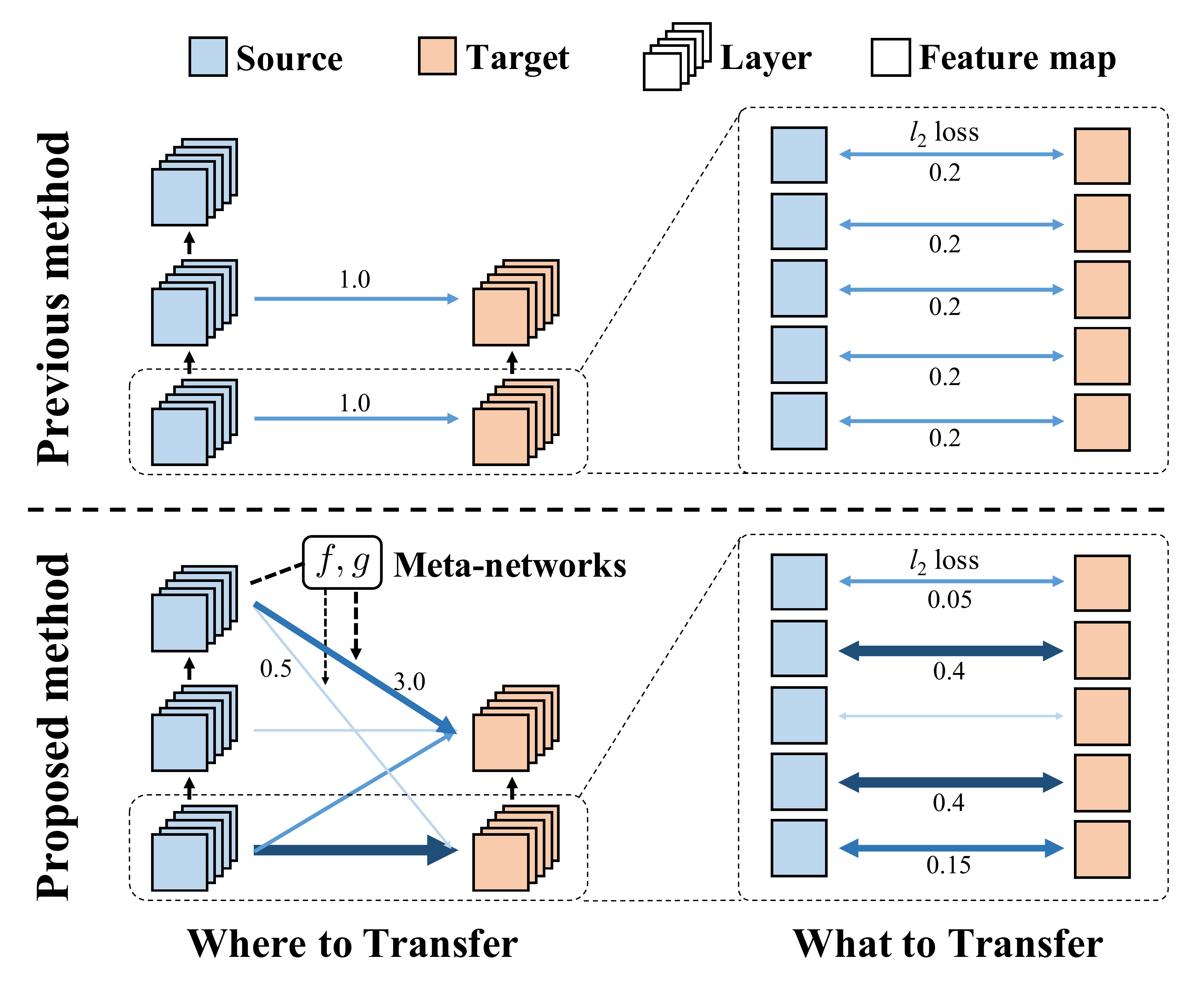}}
\vspace{-0.1in}
\caption{\textbf{Top:} Prior approaches. Knowledge transfer between two networks is done between
\textit{hand-crafted chosen pairs} of layers
\textit{without considering importance of channels}.
\textbf{Bottom:} Our meta-transfer method. The meta-networks $f,~g$ automatically decide
\textit{amounts of knowledge transfer between layers} of the two networks
and \textit{importance of channels} when transfer.
Line width indicates an amount of transfer in pairs of transferring layers
and channels.}\label{fig:overall}
\vspace{-0.4in}
\end{center}
\end{figure}

Several existing works can be applied to this challenging scenario of knowledge transfer between heterogeneous DNNs and tasks.
Learning without forgetting (LwF) \cite{li2018learning} proposes to use knowledge distillation, suggested in
Hinton et al.~\yrcite{hinton-kd-44873}, for transfer learning by introducing an additional output layer on a
target model,
and thus it can be applied to situations where the source and target tasks are different.
FitNet \cite{Romero15-iclr} proposes a teacher-student training scheme for transferring the knowledge from a wider teacher network to a thinner student network,
by using teacher's feature maps to guide the learning of the student. To guide the student network, FitNet uses $\ell_2$ matching loss between the source and target features. Attention transfer \cite{Zagoruyko2017AT} and Jacobian matching \cite{pmlr-v80-srinivas18a} suggest similar approaches to FitNet, but use attention maps generated from feature maps or Jacobians for transferring the source knowledge.

Our motivation is that these methods, while allowing to transfer knowledge between heterogeneous source and target tasks/architectures, have no mechanism to identify which source information to transfer, between which layers of the networks. Some source information is more important than others, while some are irrelevant or even harmful depending on the task difference. For example, since network layers generate representations at different level of abstractions~\cite{zeiler2014visualizing}, the information of lower layers might be more useful when the input domains of the tasks are similar, but the actual tasks are different (e.g., fine-grained image classification tasks). Furthermore, under heterogeneous network architectures, it is not straightforward to associate a layer from the source network with one from the target network. Yet, since there was no mechanism to learn what to transfer to where, existing approaches require a careful manual configuration of layer associations between the source and target networks depending on tasks, which cannot be optimal.

{\bf Contribution.}
To tackle this problem, we propose a novel transfer learning method based on the concept of \emph{meta-learning}~{\cite{naik1992meta,thrun2012learning}} that learns \emph{what} information to transfer to \emph{where}, from source networks to target networks with heterogeneous architectures and tasks. Our goal is learning to learn transfer rules for performing knowledge transfer in an automatic manner, considering the difference in the architectures and tasks between source and target, without hand-crafted tuning of transfer configurations. Specifically, we learn meta-networks that generate the weights for each feature and between each pair of source and target layers, jointly with the target network.
Thus, it can automatically learn to identify which source network knowledge is useful, and where it should transfer to (see Figure~\ref{fig:overall}). We validate our method, learning to transfer what and where (L2T-ww), to multiple source and target task combinations between heterogeneous DNN architectures, and obtain significant improvements over existing transfer learning methods.
Our contributions are as follows:
\begin{itemize}
\item We introduce meta-networks for transfer learning that automatically decide
which feature maps (channels) of a source model
are useful and relevant for learning a target task
and which source layers should be transferred to which target layers.
\item
To learn the parameters of meta-networks,
we propose an efficient meta-learning scheme.
Our main novelty is to
evaluate the one-step adaptation performance (meta-objective)
of a target model learned by  minimizing
the transfer
objective only (as an inner-objective).
This scheme significantly accelerates
the inner-loop procedure, compared to the standard scheme.
\item
The proposed method
achieves significant improvements over baseline transfer learning methods in our experiments.
For example, in the ImageNet experiment, our meta-transfer learning method achieves $65.05\%$ accuracy on CUB200,
while the second best baseline obtains $58.90\%$.
In particular,
our method outperforms baselines with a large margin when the target task has an
insufficient number of training samples and when transferring from multiple source models.
\end{itemize}

{\bf Organization.}
The rest of the paper is organized as follows.
In Section \ref{sec:method}, we describe our method for selective knowledge transfer, and training scheme for learning the proposed meta-networks.
Section \ref{sec:exp} shows our
experimental results under various settings, and Section \ref{sec:con} states the conclusion.

%% file: Methods.tex
\section{Learning What and Where to Transfer}\label{sec:method}

Our goal is to learn to transfer useful knowledge from the source network to the target network, without requiring manual layer association or feature selection. To this end, we propose a meta-learning method that learns what knowledge of the source network to transfer to which layer in the target network. In this paper, we primarily focus on transfer learning between convolutional neural networks,
but our method is generic and is applicable to other types of deep neural networks as well.

In Section \ref{sec:method:wfm}, we describe meta-networks that learn
what to transfer (for selectively transfer only the useful channels/features to a target model),
and where to transfer (for deciding a layer-matching configuration that encourages
learning a target task).
Section \ref{sec:method:training} presents how to train the proposed meta-networks jointly with the target network.

\begin{figure*}[t]
\centering
\subfigure[Where to transfer]{
\includegraphics[width=0.48\textwidth]{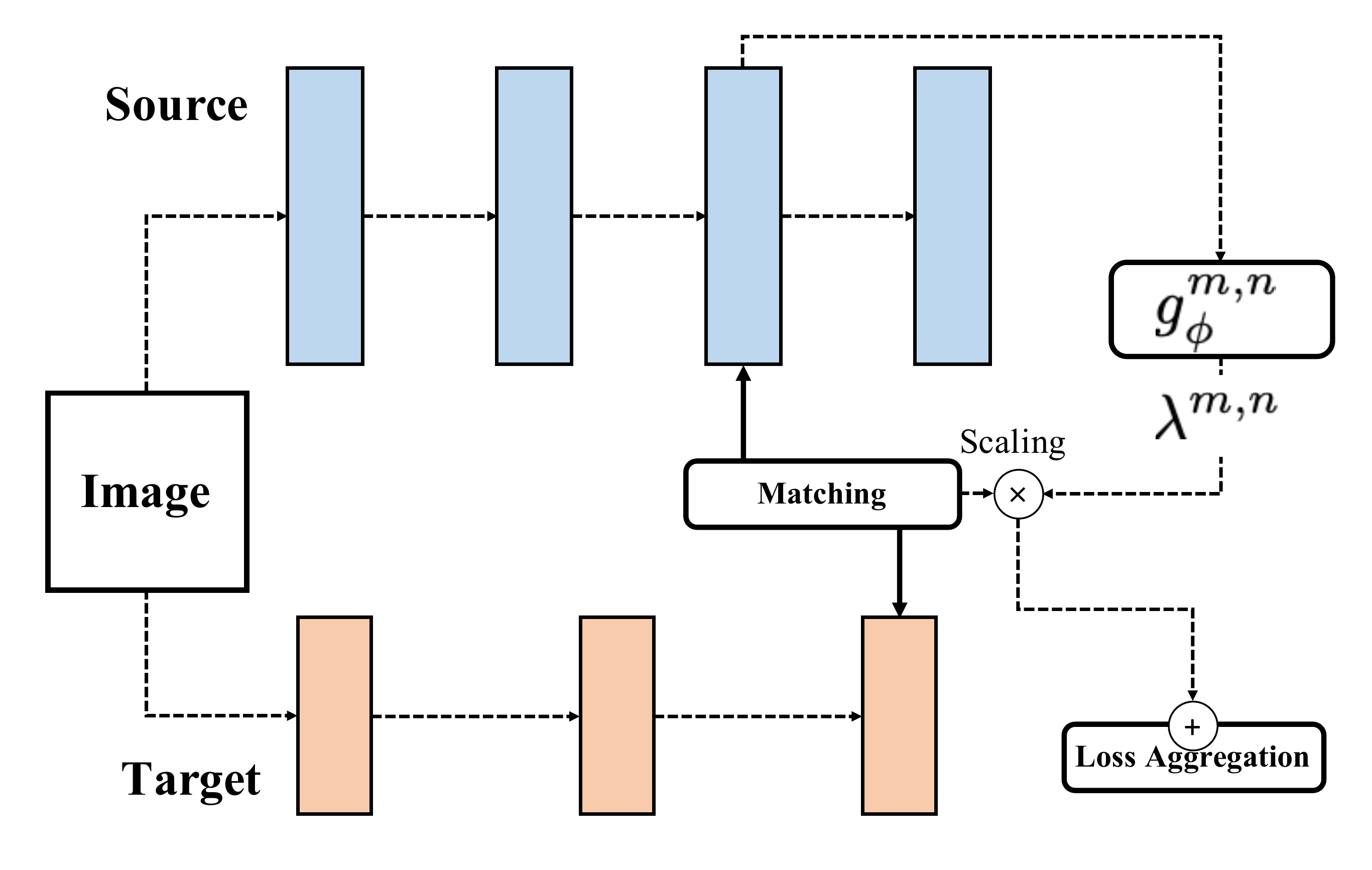}
\label{fig:method:where}}
\quad\quad
\subfigure[What to transfer]{
\includegraphics[width=0.317153748\textwidth]{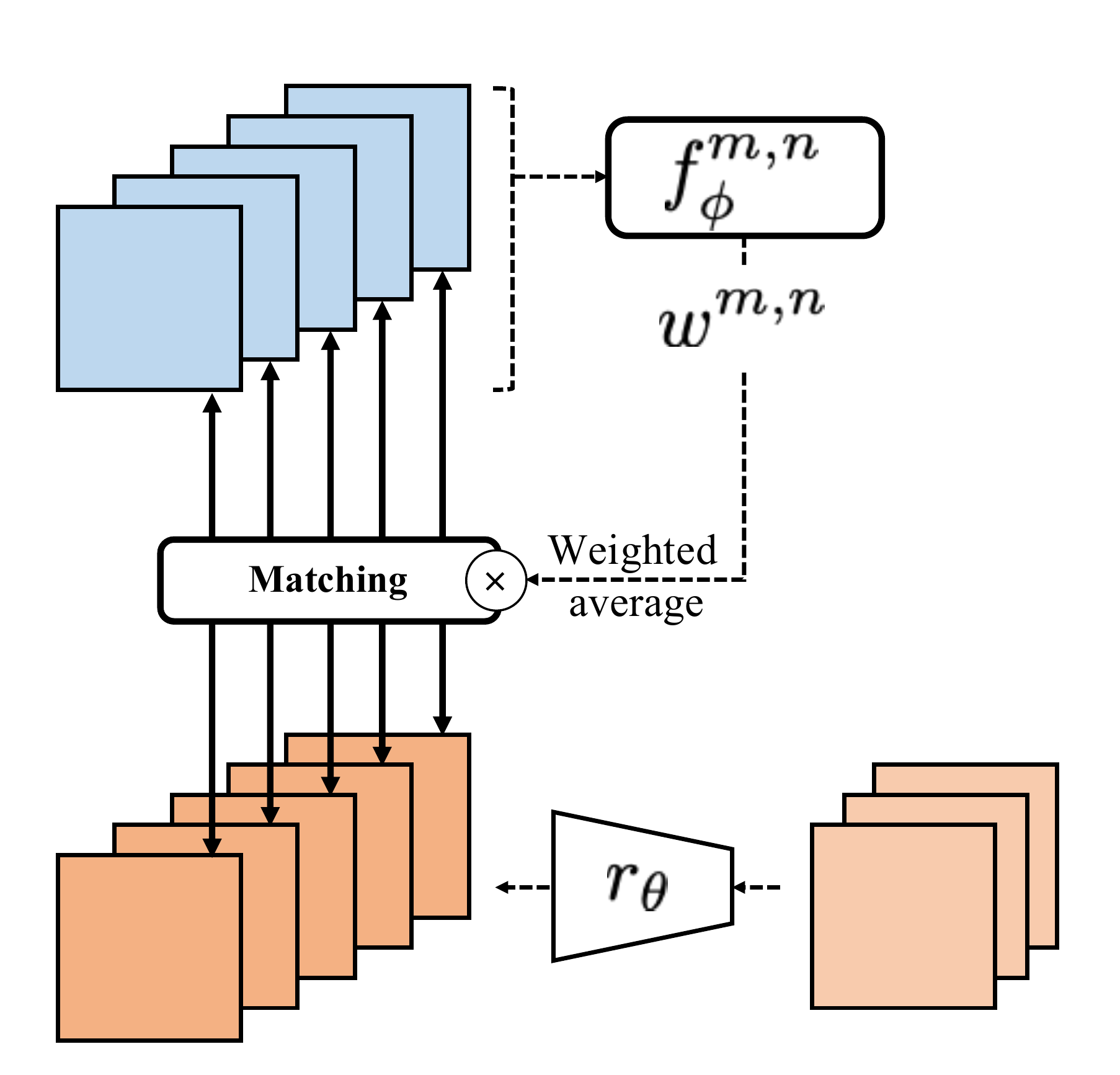}
\label{fig:method:what}}
\caption{Our meta-transfer learning method for selective knowledge transfer. The meta-transfer networks are parameterized by $\phi$ and are learned
via meta-learning.
The dashed lines indicate flows of tensors such as feature maps, and solid lines
denote $\ell_2$ feature matching.
(a) $g_\phi^{m,n}$ outputs weights of matching pairs $\lambda^{m,n}$
between
the $m^\text{th}$ and $n^\text{th}$ layers of the source and target models,
respectively, and (b) $f_\phi^{m,n}$ outputs
weights for each channel.
}\label{fig:method}
\vspace{-0.1in}
\end{figure*}

\subsection{Weighted Feature Matching}\label{sec:method:wfm}

If a convolutional neural network is well-trained on a task, then its intermediate feature spaces should have useful knowledge for the task. Thus, mimicking the well-trained features might be helpful for training another network.
To formalize the loss forcing this effect, let $x$ be an input, and $y$ be the corresponding (ground-truth) output. For image classification tasks, $\{x\}$ and $\{y\}$ are images and their class labels. Let $S^m(x)$ be
intermediate feature maps of the $m^\text{th}$ layer of the pre-trained source network $S$.
{Our goal} is then to train another target network $T_\theta$
with parameter $\theta$ utilizing the knowledge of $S$. Let $T^n_\theta(x)$ be
intermediate feature maps of the $n^\text{th}$ layer of the target network. Then, we minimize the following $\ell_2$ objective, similar to that used in FitNet \cite{Romero15-iclr},
to transfer the knowledge from  $S^m(x)$ to $T^n_\theta(x)$: 
\begin{align*}
    \lVert r_\theta(T^n_\theta(x)) - S^m(x) \rVert_2^2
\end{align*}
where $r_\theta$ is a linear transformation parameterized by $\theta$ such as a pointwise convolution.
We refer to this method as \emph{feature matching}.
Here, the parameter $\theta$ consists of both the parameter for linear-transformation $r_\theta$
and non-linear neural network $T_\theta$, where the former is only necessary in training the
latter and is not required at testing time.

\paragraph{What to transfer.}
In general transfer learning settings, the target model is trained for a task that is different from that of the source model. In this case, not all the intermediate features of the source model may be useful to learn the target task.
Thus, to give more attention on the useful channels, we consider a weighted feature matching loss that can emphasize
the channels according to their utility on the target task:
\begin{align*}
\small
    &\mathcal{L}_{\tt wfm}^{m,n}(\theta|x,w^{m,n}) \\
    &=\frac{1}{HW}\sum_c w_c^{m,n}\sum_{i,j}(r_\theta(T^n_\theta(x))_{c,i,j}-S^m(x)_{c,i,j})^2,
\end{align*}
where
$H\times W$ is the spatial size of $S^m(x)$ and $r_\theta(T_\theta^n(x))$,
the inner-summation is over $i\in \{1,2,\dots H\}$ and $j\in\{1,2,\dots W\}$,
and $w_c^{m,n}$ is the non-negative weight of channel $c$ with $\sum_c w_c^{m,n}=1$.
Since the important channels to transfer can vary for each input image, we set channel weights as a function, $w^{m,n}=[w^{m,n}_c]=f^{m,n}_\phi(S^m(x))$, by taking the softmax output of a small meta-network which takes features of source models as an input.
We let $\phi$ {denote} the parameters of meta-networks throughout this paper.

\paragraph{Where to transfer.} When transferring knowledge from a source model to a target model, deciding pairs $(m,n)$ of layers in the source and target model is crucial to its effectiveness.
Previous approaches~\citep{Romero15-iclr,Zagoruyko2017AT} select
the pairs manually based on prior knowledge
of architectures or semantic similarities between tasks.
For example, attention transfer \citep{Zagoruyko2017AT}
matches the last feature maps of each group of residual blocks
in ResNet~\citep{he2016deep}.
However, finding the optimal layer association is not a trivial problem and requires exhaustive tuning based on trial-and-error, given models with different numbers of layers or heterogeneous architectures, e.g., between ResNet \citep{he2016deep} and VGG \citep{simonyan2014very}.
Hence, we introduce a learnable parameter
$\lambda^{m,n}\ge 0$ for each pair $(m,n)$ which
can decide the amount of transfer between the $m^\text{th}$ and $n^\text{th}$
layers of source and target models, respectively.
We also set $\lambda^{m,n}=g^{m,n}_\phi(S^m(x))$
for each pair $(m,n)$ as an output of a meta-network $g^{m,n}$
that automatically decides important pairs of layers
for learning the target task.

The combined transfer loss
given the weights of channels $w$ and weights of matching pairs $\lambda$ is
$$\mathcal{L}_{\tt wfm}(\theta|x,\phi)  =
    \sum_{(m,n)\in\mathcal{C}}\lambda^{m,n}\mathcal{L}_{\tt wfm}^{m,n}(\theta|x,w^{m,n}),$$
    where $\mathcal{C}$ be a set of candidate pairs.
Our final loss $\mathcal{L}_{\tt total}$ to train a target model then is given as:
\begin{align*}
    \mathcal{L}_{\tt total}(\theta|x,y,\phi) & =\mathcal{L}_{\tt org}(\theta|x,y) + \beta\mathcal{L}_{\tt wfm}(\theta|x,\phi).
\end{align*}
where $\mathcal{L}_{\tt org}$ is the original loss (e.g., cross entropy) and $\beta>0$ is a hyper-parameter.
We note that $w^{m,n}$ and $\lambda^{m,n}$ decide what and where to transfer, respectively.
We provide an illustration of our transfer learning scheme in Figure \ref{fig:method}.

\subsection{Training Meta-Networks and Target Model}\label{sec:method:training}

Our goal is to achieve high performance on the target task when the target model is learned using the training objective $\mathcal{L}_{\tt total}(\cdot|x,y,\phi)$.
To maximize the performance, the feature matching term $\mathcal{L}_{\tt wfm}(\cdot|x,\phi)$
should encourage learning of useful features for the target task, e.g., predicting labels.
To measure and increase usefulness of the feature matching decided by meta-networks parameterized by $\phi$,
a standard approach is to use the following bilevel scheme~\citep{colson2007overview} to train $\phi$, e.g., see \cite{finn2017model, franceschi2018bilevel}:
\begin{enumerate}
    \item Update $\theta$ to minimize $\mathcal{L}_{\tt total}(\theta|x,y,\phi)$ for $T$ times.
    \item Measure $\mathcal{L}_{\tt org}(\theta|x,y)$ and update $\phi$ to minimize it.
\end{enumerate}
In the above,
the actual objective  $\mathcal{L}_{\tt total}$
for learning the target model
is used in the inner-loop,
and
the original loss $\mathcal{L}_{\tt org}$ is used as a meta-objective to measure
the effectiveness of $\mathcal{L}_{\tt total}$ for learning the target model to perform well.

However,
since our meta-networks affect the learning procedure of the target model weakly
through the regularization term $\mathcal{L}_{\tt wfm}$, their influence on
$\mathcal{L}_{\tt org}$ can be very marginal, unless one uses a very large number of
inner-loop iterations $T$.
Consequently, it
causes difficulties on updating $\phi$ using gradient $\nabla_{\phi}\mathcal{L}_{\tt org}$.
To tackle this challenge,
we propose the following alternative scheme:
\begin{enumerate}
    \item Update $\theta$ to minimize $\mathcal{L}_{\tt wfm}(\theta|x,\phi)$ for $T$ times.
    \item Update $\theta$ to minimize $\mathcal{L}_{\tt org}(\theta|x,y)$ once.
    \item Measure $\mathcal{L}_{\tt org}(\theta|x,y)$ and update $\phi$ to minimize it.
\end{enumerate}
In the first stage,
given the current parameter $\theta_0=\theta$,
we
update the target model
for $T$ times via gradient-based algorithms for minimizing
$\mathcal{L}_{\tt wfm}$.
Namely, the resulting parameter $\theta_{T}$ is learned only
using the knowledge of the source model.
Since transfer is done by the form
of feature matching,
it is feasible to train useful features for the target task by selectively mimic the source features.
More importantly, it increases the influence of the regularization term $\mathcal{L}_{\tt wfm}$
on the learning procedure of the target model in the inner-loop, since the target features
are solely trained by the source knowledge (without target labels).
The second stage is an one-step adaptation $\theta_{T+1}$
from 
$\theta_{T}$ toward the target label.
Then, in the third stage,
the task-specific objective $\mathcal{L}_{\tt org}(\theta_{T+1})$
can measure how quickly the target model has adapted (via only one step from $\theta_{T}$)
to the target task,
under the sample used in the first and second stage.
Finally, the meta-parameter $\phi$ can be trained by minimizing
$\mathcal{L}_{\tt org}(\theta_{T+1})$.
The above 3-stage scheme encourages significantly faster training of $\phi$,
compared the standard 2-stage one.
{This is because the former measures the effect of the regularization term $\mathcal{L}_{\tt wfm}$
more directly to the original $\mathcal{L}_{\tt org}$, and}
allows to choose a small $T$
to update $\phi$ meaningfully (we choose $T=2$ in our experiments).

In the case of using the vanilla gradient descent algorithm for updates,
the 3-stage training scheme to
learn meta-parameters $\phi$ can be formally written as
the following optimization task:
\begin{equation*}
\begin{aligned}
& \underset{\phi}{\text{minimize}}
& & \mathcal{L}_{\tt org}(\theta_{T+1}|x,y) \\
& \text{subject to}
& & \theta_{T+1} = \theta_T-\alpha\nabla_\theta\mathcal{L}_{\tt org}(\theta_T|x,y), \\
& & & \theta_{t+1} = \theta_t-\alpha\nabla_\theta\mathcal{L}_{\tt wfm}(\theta_t|x,\phi), \\
& & & t=0,\ldots,T-1,
\end{aligned}
\end{equation*}
where $\alpha>0$ is a learning rate.
To solve the above optimization problem,
we use Reverse-HG \citep{pmlr-v70-franceschi17a} that can compute
$\nabla_\phi\mathcal{L}_{\tt org}(\theta_{T+1}|x,y)$ efficiently using Hessian-vector products.

To train the target model jointly with meta-networks, we alternatively update the
target model parameters $\theta$ and the meta-network parameters $\phi$. We first update the target model for a single step with objective
$\mathcal{L}_{\tt total}(\theta|x, y, \phi)$.
Then, given current target model parameters, we update the meta-networks parameters $\phi$
using the 3-stage bilevel training scheme described above.
This eliminates
an additional meta-training phase for learning $\phi$.
The proposed training scheme is formally outlined in Algorithm~\ref{alg:alg}.

\begin{algorithm}[tb]
   \caption{Learning of $\theta$ with meta-parameters $\phi$}
   \label{alg:alg}
\begin{algorithmic}
   \STATE {\bfseries Input:} Dataset $\mathcal{D}_{\tt train}=\{(x_i,y_i)\}$, learning rate $\alpha$
   \REPEAT
   \STATE Sample a batch $\mathcal{B}\subset\mathcal{D}_{\tt train}$ with $|\mathcal{B}|=B$
   \STATE Update $\theta$ to minimize $\frac{1}{B}\sum_{(x,y)\in\mathcal{B}}\mathcal{L}_{\tt total}(\theta|x,y,\phi)$
   \STATE Initialize $\theta_0\leftarrow\theta$
   \FOR{$t=0$ {\bfseries to} $T-1$}
   \STATE $\theta_{t+1}\leftarrow\theta_t-\alpha\nabla_\theta\frac{1}{B}\sum_{(x,y)\in\mathcal{B}}\mathcal{L}_{\tt wfm}(\theta_t|x,\phi)$
   \ENDFOR
   \STATE $\theta_{T+1}\leftarrow\theta_T-\alpha\nabla_\theta\frac{1}{B}\sum_{(x,y)\in\mathcal{B}}\mathcal{L}_{\tt org}(\theta_T|x,y)$
   \STATE Update $\phi$ using $\nabla_\phi\frac{1}{B}\sum_{(x,y)\in\mathcal{B}}\mathcal{L}_{\tt org}(\theta_{T+1}|x,y)$
   \UNTIL{done}
\end{algorithmic}
\end{algorithm}

%% file: Experiments.tex
\section{Experiments}\label{sec:exp}

\begin{figure*}[t]
\centering
\subfigure[Single]{
\includegraphics[width=0.235\textwidth]{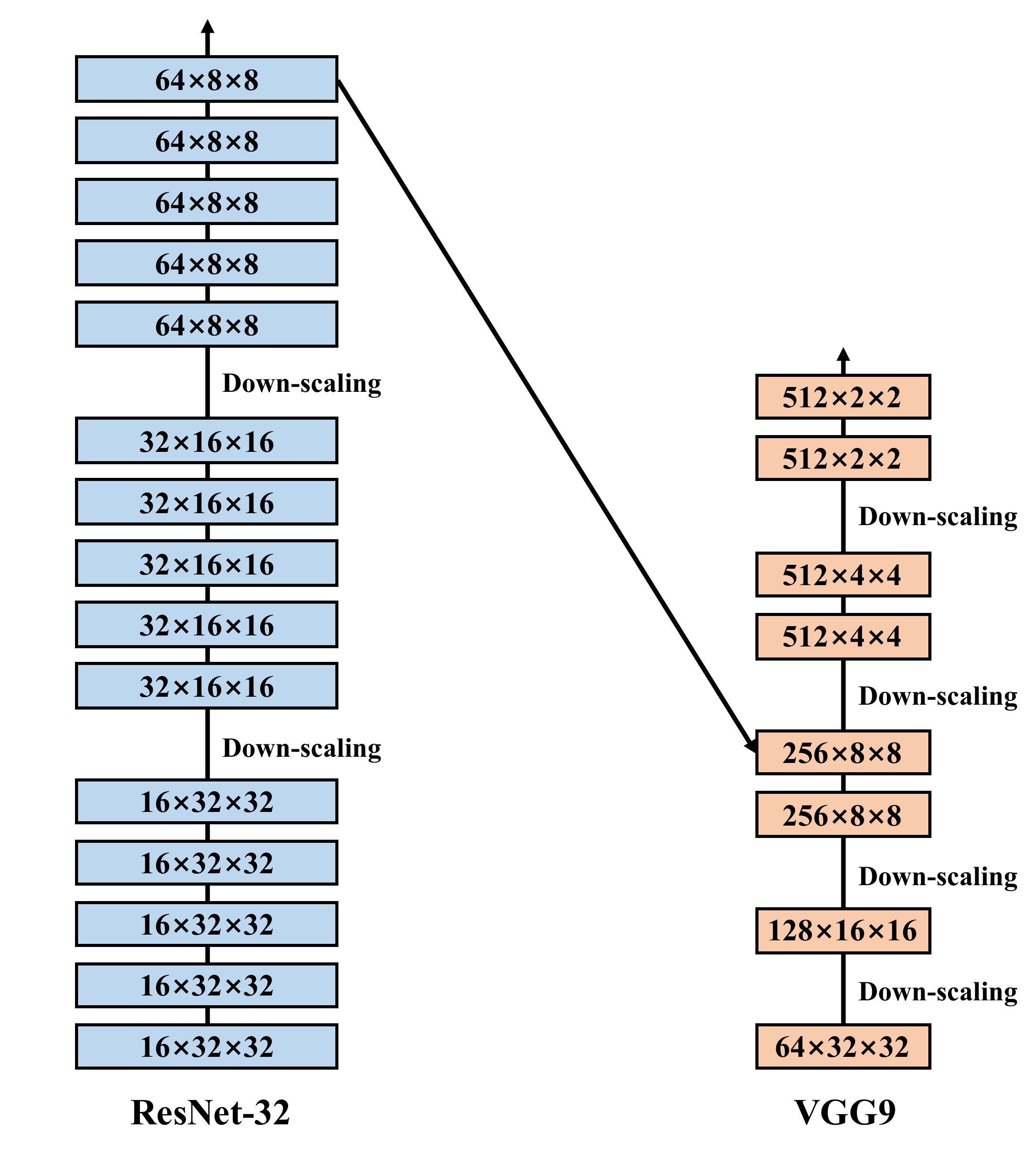}
\label{fig:resvgg-last}}
\subfigure[One-to-one]{
\includegraphics[width=0.235\textwidth]{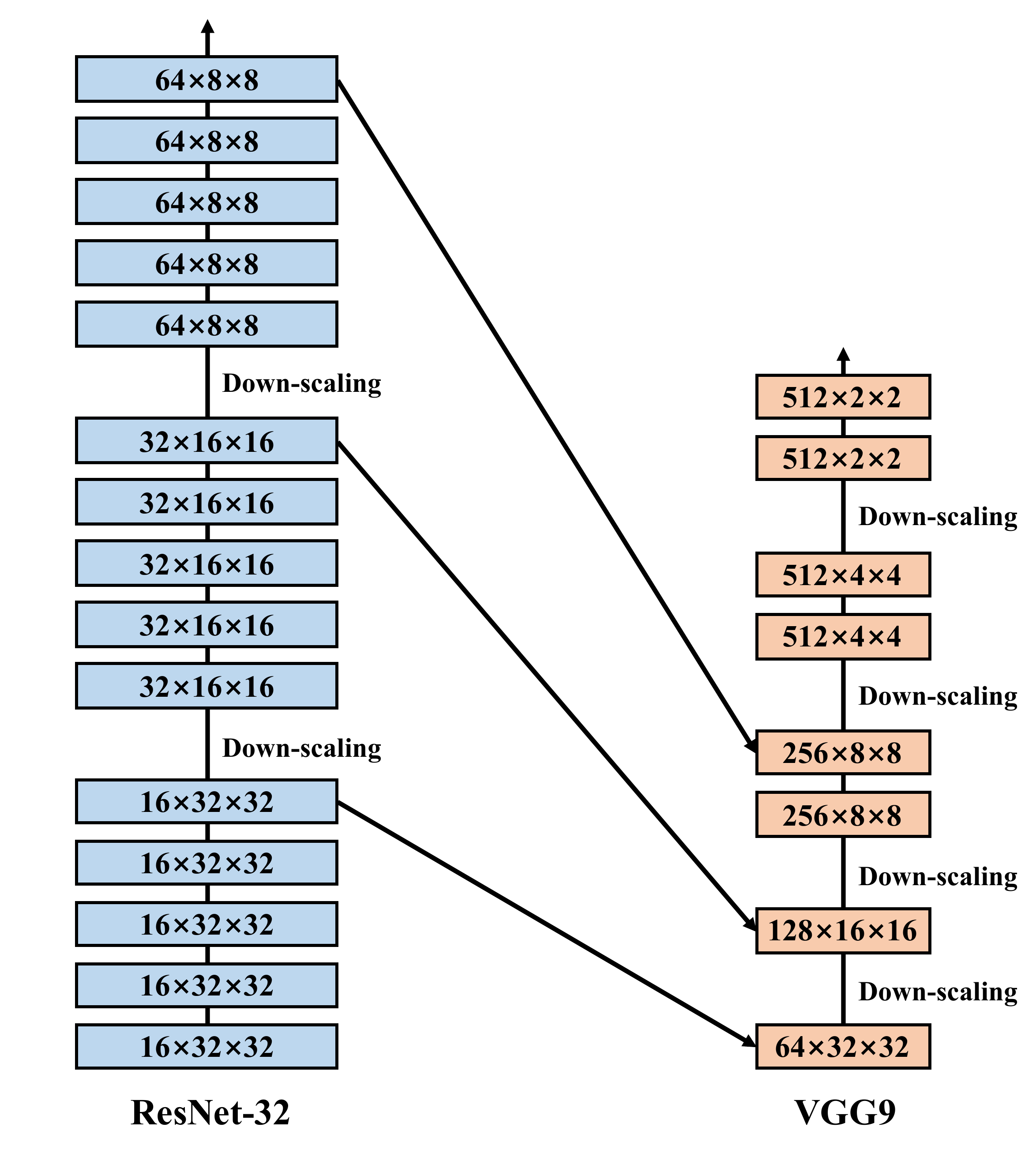}
\label{fig:resvgg-parallel}}
\subfigure[All-to-all]{
\includegraphics[width=0.235\textwidth]{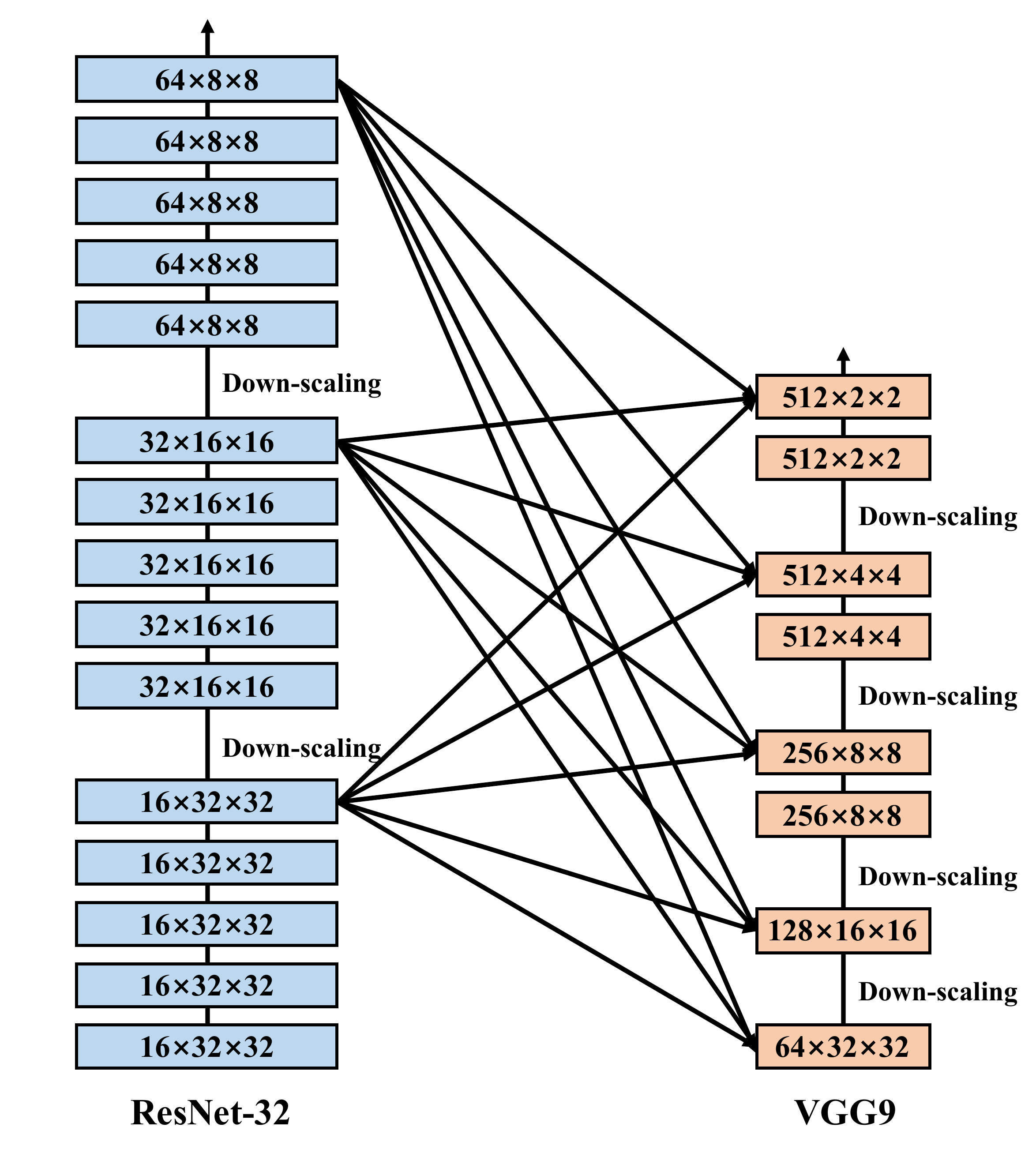}
\label{fig:resvgg-all-before}}
\subfigure[Learned matching]{
\includegraphics[width=0.235\textwidth]{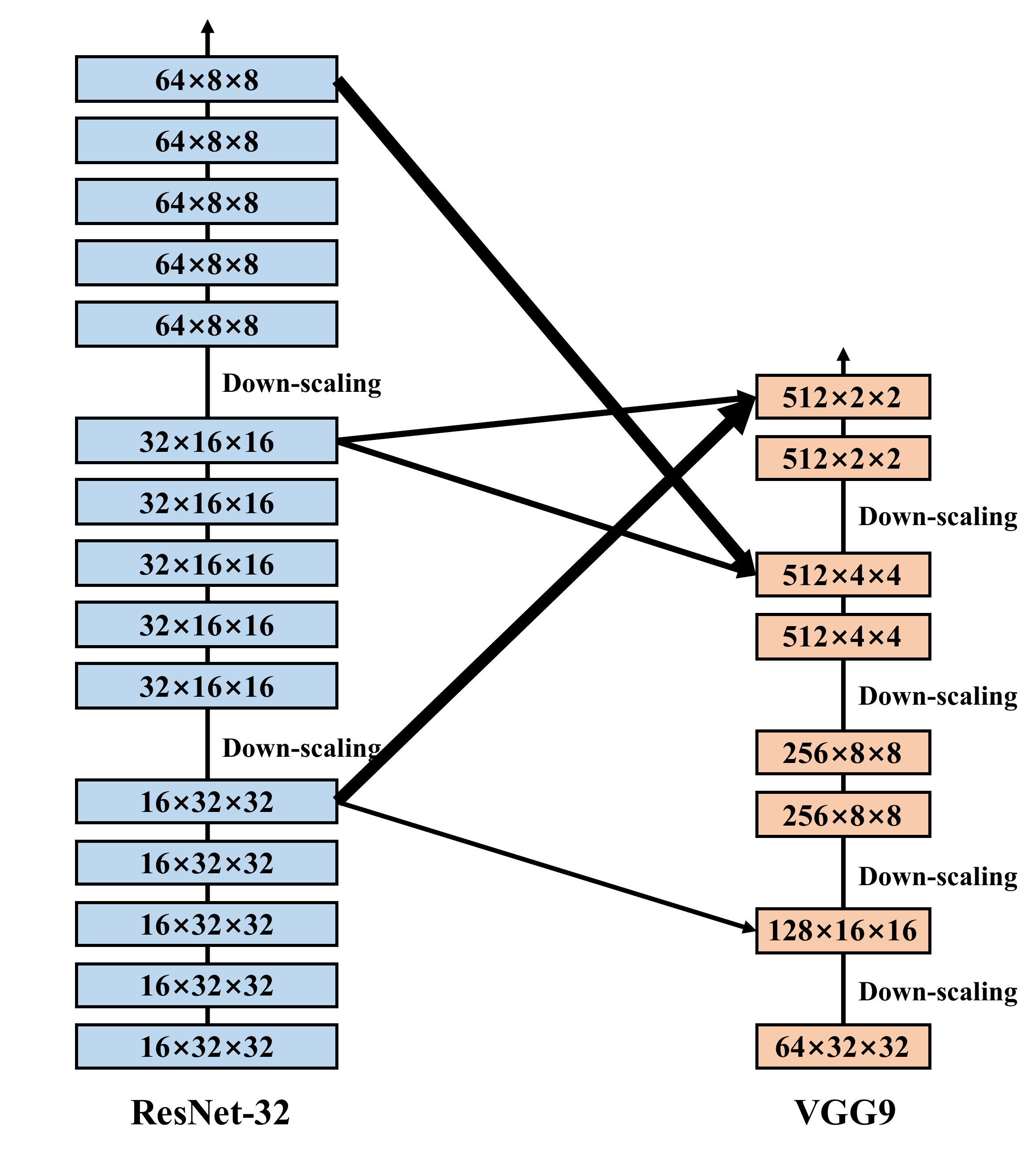}
\label{fig:resvgg-all}}
\caption{
(a)-(c) Matching configurations $\mathcal{C}$ between ResNet32 (left) and VGG9 (right).
(d)
The amount $\lambda^{m,n}$ of transfer between layers after learning.
Line widths indicates the transfer amount. We omit the lines when $\lambda^{m,n}$ is less than $0.1$.
}\label{fig:resvgg}
\end{figure*}

We validate our meta-transfer learning method that learns what and where to transfer, between heterogeneous network architectures and tasks.

\subsection{Setups}\label{sec:exp:setup}
\paragraph{Network architectures and tasks for source and target.}
To evaluate various transfer learning methods including ours,
we perform experiments on two scales of image classification tasks, $32\times32$ and $224\times224$.
For $32\times32$ scale, we use the  TinyImageNet\footnote{https://tiny-imagenet.herokuapp.com/}
dataset as a source task, and CIFAR-10, CIFAR-100~\citep{krizhevsky2009learning}, and STL-10~\citep{coates2011analysis} datasets as target tasks.
We train 32-layer ResNet \citep{he2016deep} and 9-layer VGG \citep{simonyan2014very} on the source and target tasks, respectively.
For $224\times224$ scale, the ImageNet~\citep{deng2009imagenet} dataset is used as
a source dataset, and Caltech-UCSD Bird 200~\citep{WahCUB_200_2011}, MIT Indoor Scene Recognition~\citep{quattoni2009recognizing},
Stanford 40 Actions~\citep{yao2011human}
and Stanford Dogs~\citep{stanford_dogs} datasets as target tasks.
For these datasets, we
use 34-layer and 18-layer ResNet as a source and target model, respectively, unless
otherwise stated.

\paragraph{Meta-network architecture.}
For all experiments, we construct the meta-networks as $1$-layer fully-connected networks
for each pair $(m,n)\in\mathcal{C}$
where $\mathcal{C}$ is the set of candidates of pairs, or {\it matching configuration}
(see Figure \ref{fig:resvgg}).
It takes the globally average pooled features of the $m^\text{th}$ layer of the source network as an input,
and outputs $w_c^{m,n}$ and $\lambda^{m,n}$.
As for the channel assignments $w$, we use the softmax activation to generate them while satisfying $\sum_cw_c^{m,n}=1$, and for transfer amount $\lambda$ between layers,
we commonly use ReLU6 \citep{krizhevsky2010convolutional}, $\max(0,\min(6,x))$ to ensure non-negativeness of $\lambda$ and to prevent $\lambda^{m,n}$ from becoming too large.

\begin{figure*}[t]
\centering
\subfigure[Transfer from $S^{1}(x)$]{
\includegraphics[width=0.31\textwidth]{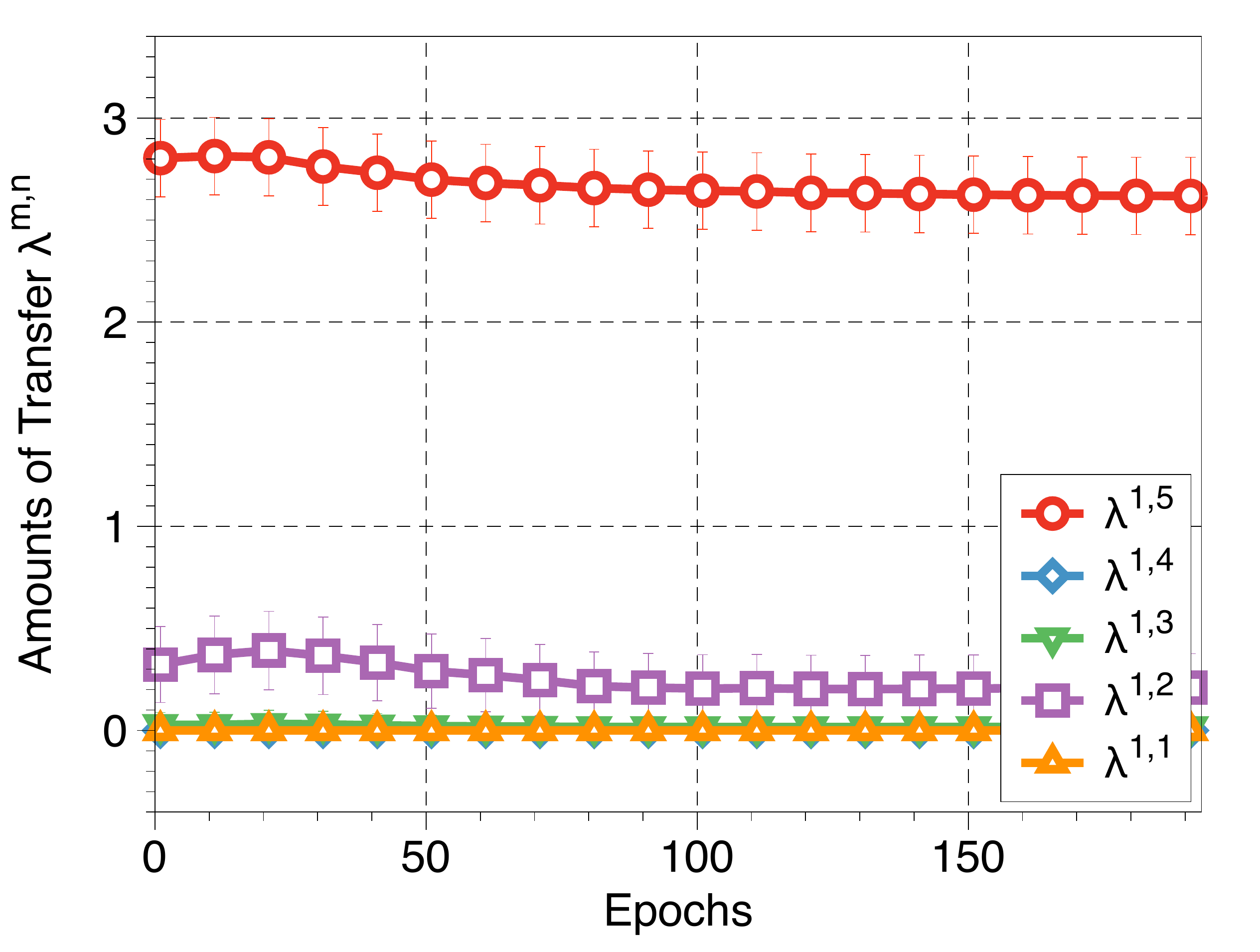}
\label{fig:1st_lw}}
\subfigure[Transfer from $S^{2}(x)$]{
\includegraphics[width=0.31\textwidth]{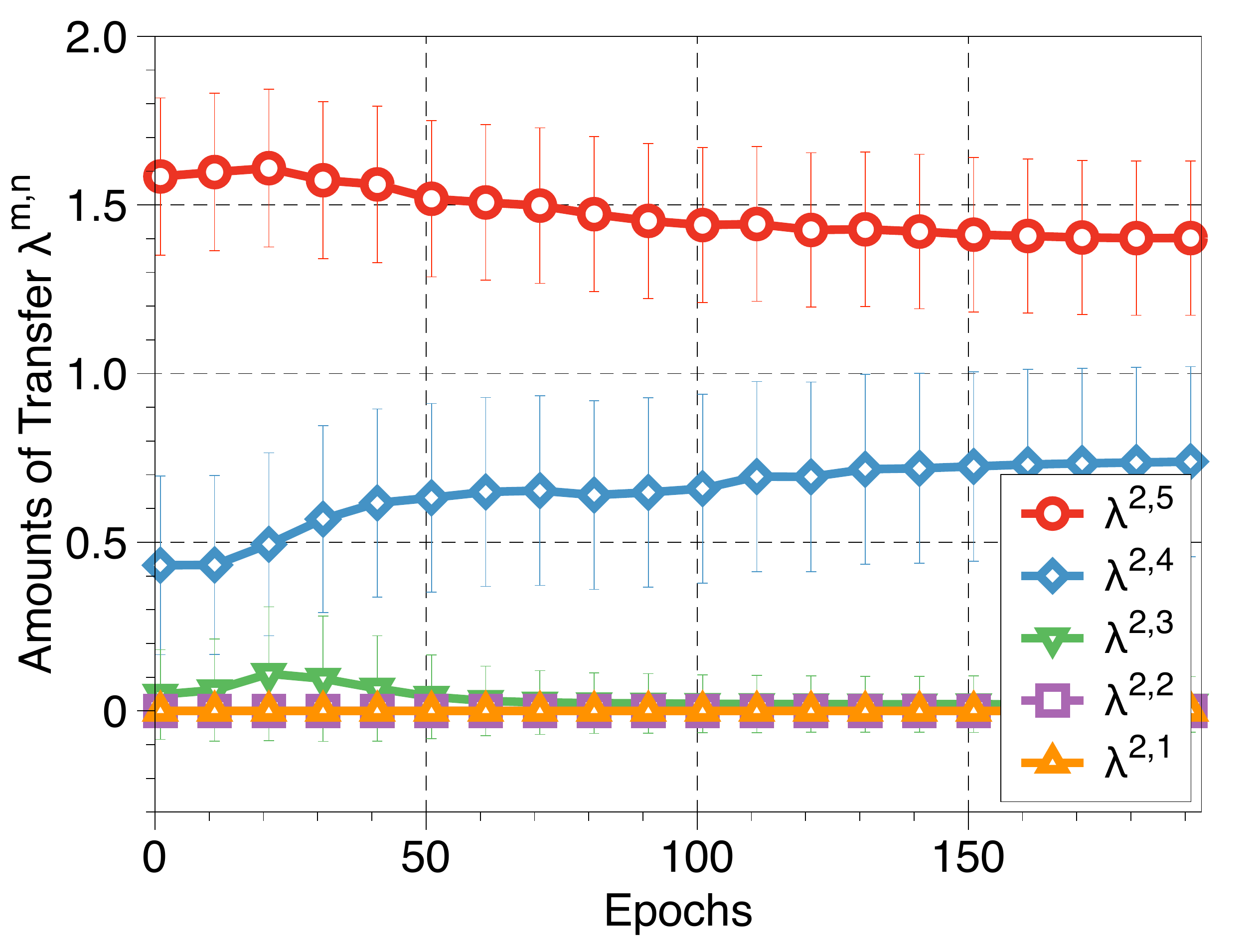}
\label{fig:2nd_lw}}
\subfigure[Transfer from $S^{3}(x)$]{
\includegraphics[width=0.31\textwidth]{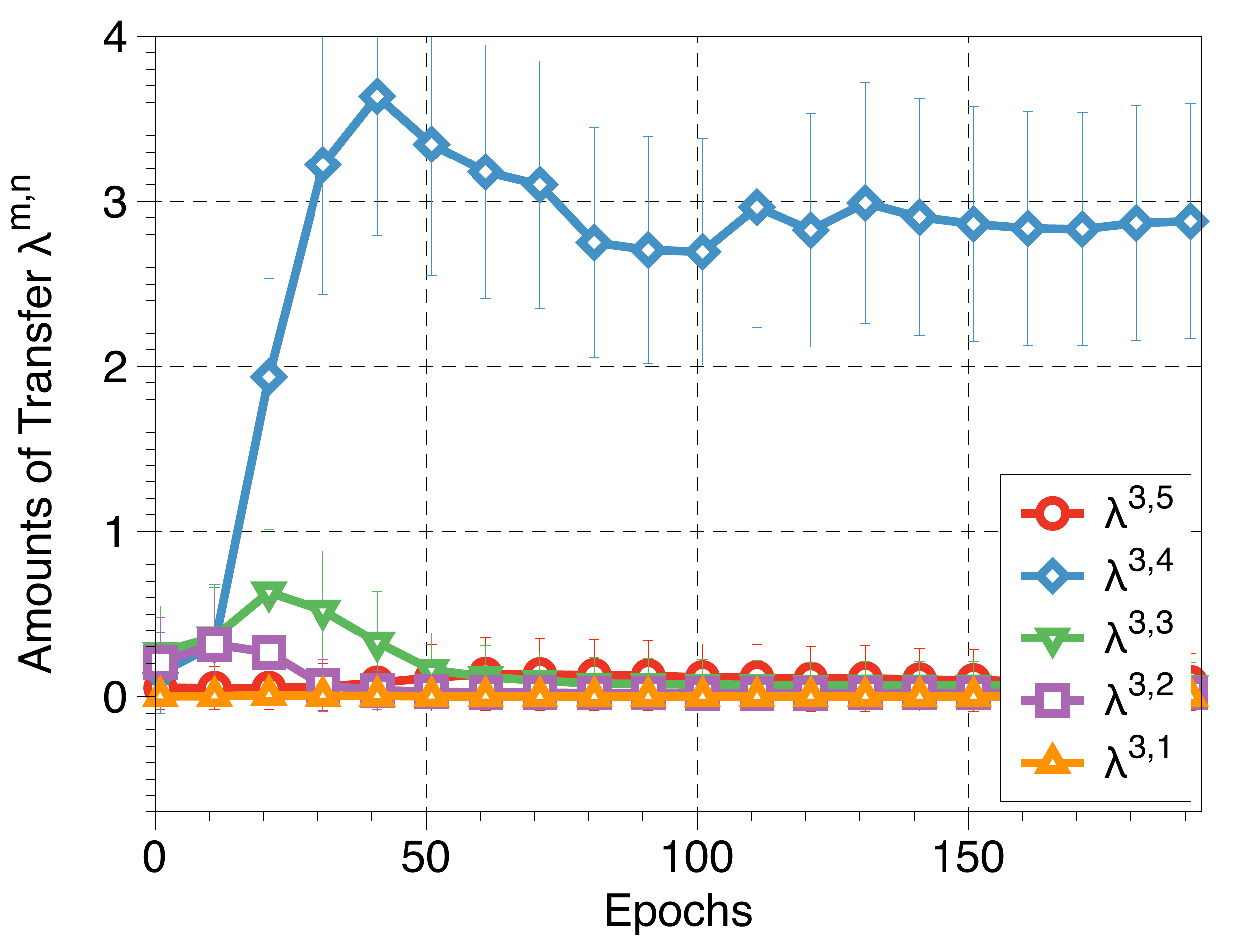}
\label{fig:3rd_lw}}
\caption{\small Change of $\lambda^{m,n}$ during training for STL-10 as the targe task with
TinyImageNet as the source task. We plot mean and standard deviation of $\lambda^{m,n}$ of all
samples for every 10 epochs.}\label{fig:lw}
\end{figure*}

\begin{table*}[t]
  \centering
  \caption{Classification accuracy (\%) of transfer learning from TinyImageNet ($32\times32$) or ImageNet
  ($224\times224)$ to CIFAR-100, STL-10, Caltech-UCSD Bird 200 (CUB200), MIT Indoor Scene Recognition (MIT67), Stanford 40 Actions (Stanford40) and Stanford Dogs datasets.
  For TinyImageNet, ResNet32 and VGG9 are used as a source and target model, respectively, and
  ResNet34 and ResNet18 are used for ImageNet.
  }\label{tbl:tiny}
  \vspace{0.1in}
  \begin{tabular}{ccccccc}
  \toprule
  Source task          & \multicolumn{2}{c}{TinyImageNet} & \multicolumn{4}{c}{ImageNet}      \\ \cmidrule(lr){1-1}\cmidrule(lr){2-3}\cmidrule(lr){4-7}
  Target task          & CIFAR-100             & STL-10                & CUB200                & MIT67                 & Stanford40            & Stanford Dogs         \\ \midrule
  Scratch              & \ms{67.69}{0.22}      & \ms{65.18}{0.91}      & \ms{42.15}{0.75}      & \ms{48.91}{0.53}      & \ms{36.93}{0.68}      & \ms{58.08}{0.26}      \\
  LwF                  & \ms{69.23}{0.09}      & \ms{68.64}{0.58}      & \ms{45.52}{0.66}      & \ms{53.73}{2.14}      & \ms{39.73}{1.63}      & \ms{66.33}{0.45}      \\
  AT     (one-to-one)  & \ms{67.54}{0.40}      & \ms{74.19}{0.22}      & \ms{57.74}{1.17}      & \ms{59.18}{1.57}      & \ms{59.29}{0.91}      & \ms{69.70}{0.08}      \\
  LwF+AT (one-to-one)  & \ms{68.75}{0.09}      & \ms{75.06}{0.57}      & \ms{58.90}{1.32}      & \ms{61.42}{1.68}      & \ms{60.20}{1.34}      & \ms{72.67}{0.26}       \\
  FM     (single)      & \ms{69.40}{0.67}      & \ms{75.00}{0.34}      & \ms{47.60}{0.31}      & \ms{55.15}{0.93}      & \ms{42.93}{1.48}      & \ms{66.05}{0.76}      \\
  FM     (one-to-one)  & \ms{69.97}{0.24}      & \ms{76.38}{1.18}      & \ms{48.93}{0.40}      & \ms{54.88}{1.24}      & \ms{44.50}{0.96}      & \ms{67.25}{0.88}      \\ \midrule
  L2T-w  (single)      & \ms{70.27}{0.09}      & \ms{74.35}{0.92}      & \ms{51.95}{0.83}      & \ms{60.41}{0.37}      & \ms{46.25}{3.66}      & \ms{69.16}{0.70}      \\
  L2T-w  (one-to-one)  & \ms{70.02}{0.19}      & \ms{76.42}{0.52}      & \ms{56.61}{0.20}      & \ms{59.78}{1.90}      & \ms{48.19}{1.42}      & \ms{69.84}{1.45}      \\
  L2T-ww (all-to-all)  & \bf{\ms{70.96}{0.61}} & \bf{\ms{78.31}{0.21}} & \bf{\ms{65.05}{1.19}} & \bf{\ms{64.85}{2.75}} & \bf{\ms{63.08}{0.88}} & \bf{\ms{78.08}{0.96}} \\
  \bottomrule
  \end{tabular}
\end{table*}

\paragraph{Compared schemes for transfer learning.}
We compare our methods with the following prior methods and their combinations: learning without forgetting (LwF)~\cite{li2018learning},
attention transfer (AT)~\cite{Zagoruyko2017AT} and
unweighted feature matching (FM)~\cite{Romero15-iclr}.\footnote{{
In our experimental setup, we reproduce similar
relative improvements from the scratch for these baselines as reported in the original papers.
We do not report the results of
Jacobian matching (JM)~\cite{pmlr-v80-srinivas18a} as
the improvement of LwF+AT+JM over LwF+AT is marginal in our setups.
}}
Here, AT and FM transfer knowledge on feature-level as like ours
by matching attention maps or feature maps between source and target layers, respectively.
The feature-level transfer methods
generally choose
layers just before down-scaling, e.g.,
the last layer of each residual group for ResNet,
and match pairs of the layers
of same spatial size.
Following this convention, we evaluate two hand-crafted configurations (single, one-to-one)
for prior methods
and a new configurations (all-to-all) for our methods:
(a) {\it single}: use a pair of the last feature in the source model
and a layer with the same spatial size in the target model,
(b) {\it one-to-one}: connect
each layer just before down-scaling in the source model to a target layer of the same spatial size,
(c) {\it all-to-all}: use all pairs of layers just before
down-scaling, e.g., between ResNet and VGG architectures,
we consider $3\times5=15$ pairs.
For matching features of different spatial sizes, we simply use a bilinear interpolation.
These configurations are illustrated in Figure \ref{fig:resvgg}.
Among various combinations between prior methods and matching configurations,
we only report the results of those achieving the meaningful performance gains.

\begin{figure}[t]
\begin{center}
\centerline{\includegraphics[width=0.8\columnwidth]{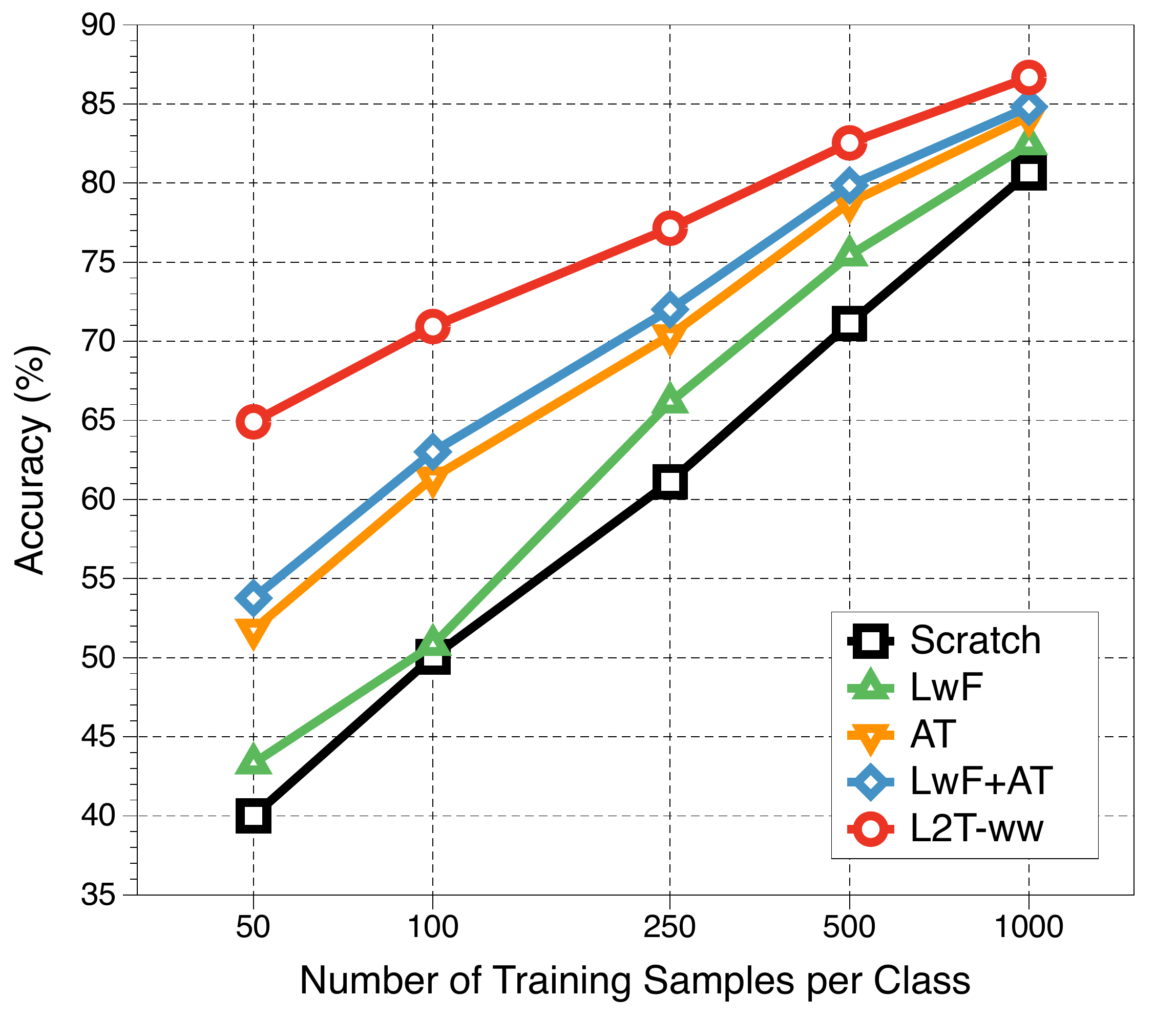}}
\caption{Transfer from TinyImageNet to CIFAR-10 with varying numbers of training samples per class in CIFAR-10.
$x$-axis is plotted in logarithmic scale.}\label{fig:fewshot}
\vspace{-0.2in}
\end{center}
\end{figure}

\afterpage{
\begin{table*}[h]
  \centering
  \caption{Classification accuracy (\%) of VGG9 on STL-10 transferred from multiple source models.
  The first source model is ResNet32 trained on TinyImageNet. The additional source model is one of three:
  ResNet20 trained on TinyImageNet, another ResNet32 trained on TinyImageNet, and ResNet32 trained on CIFAR-10.
  We report the performance of the target model transferred
  from a single source model and
  two source models.}\label{tbl:multisource-v2}
  \vspace{0.1in}
  \begin{tabular}{ccccc}
  \toprule
  First source      & \multicolumn{4}{c}{TinyImageNet~(ResNet32)}   \\ \cmidrule(lr){1-1}\cmidrule(lr){2-5}
  \multirow{1}{*}{Second source} &   \multirow{1}{*}{None} & TinyImageNet (ResNet20) & TinyImageNet (ResNet32)   & CIFAR-10 (ResNet32) \\
  \midrule
  Scratch                 & \ms{65.18}{0.91}        & \ms{65.18}{0.91}      & \ms{65.18}{0.91}         & \ms{65.18}{0.91}         \\
  LwF                     & {\ms{68.64}{0.58}}      & {\ms{68.56}{2.24}}    & {\ms{68.05}{2.12}}       & {\ms{69.51}{0.63}}       \\
  AT                      & \ms{74.19}{0.22}        & {\ms{73.24}{0.12}}    & {\ms{73.78}{1.16}}       & {\ms{73.99}{0.51}}       \\
  LwF+AT                  & \ms{75.06}{0.57}        & {\ms{74.72}{0.46}}    & {\ms{74.77}{0.30}}       & {\ms{74.41}{1.51}}       \\
  FM (single)             & {\ms{75.00}{0.34}}      & {\ms{75.83}{0.56}}    & {\ms{75.99}{0.11}}       & {\ms{74.60}{0.73}}       \\
  FM (one-to-one)         & {\ms{76.38}{1.18}}      & {\ms{77.45}{0.48}}    & {\ms{77.69}{0.79}}       & {\ms{77.15}{0.41}}       \\
  \midrule
  L2T-ww (all-to-all) & \textbf{\ms{78.31}{0.21}} & \bf{\ms{79.35}{0.41}} & \bf{\ms{79.80}{0.52}}      & \bf{\ms{80.52}{0.29}} \\
  \bottomrule
  \end{tabular}
\end{table*}
\begin{figure*}[t]
\centering
\subfigure[CUB200]{
\includegraphics[width=0.42\textwidth]{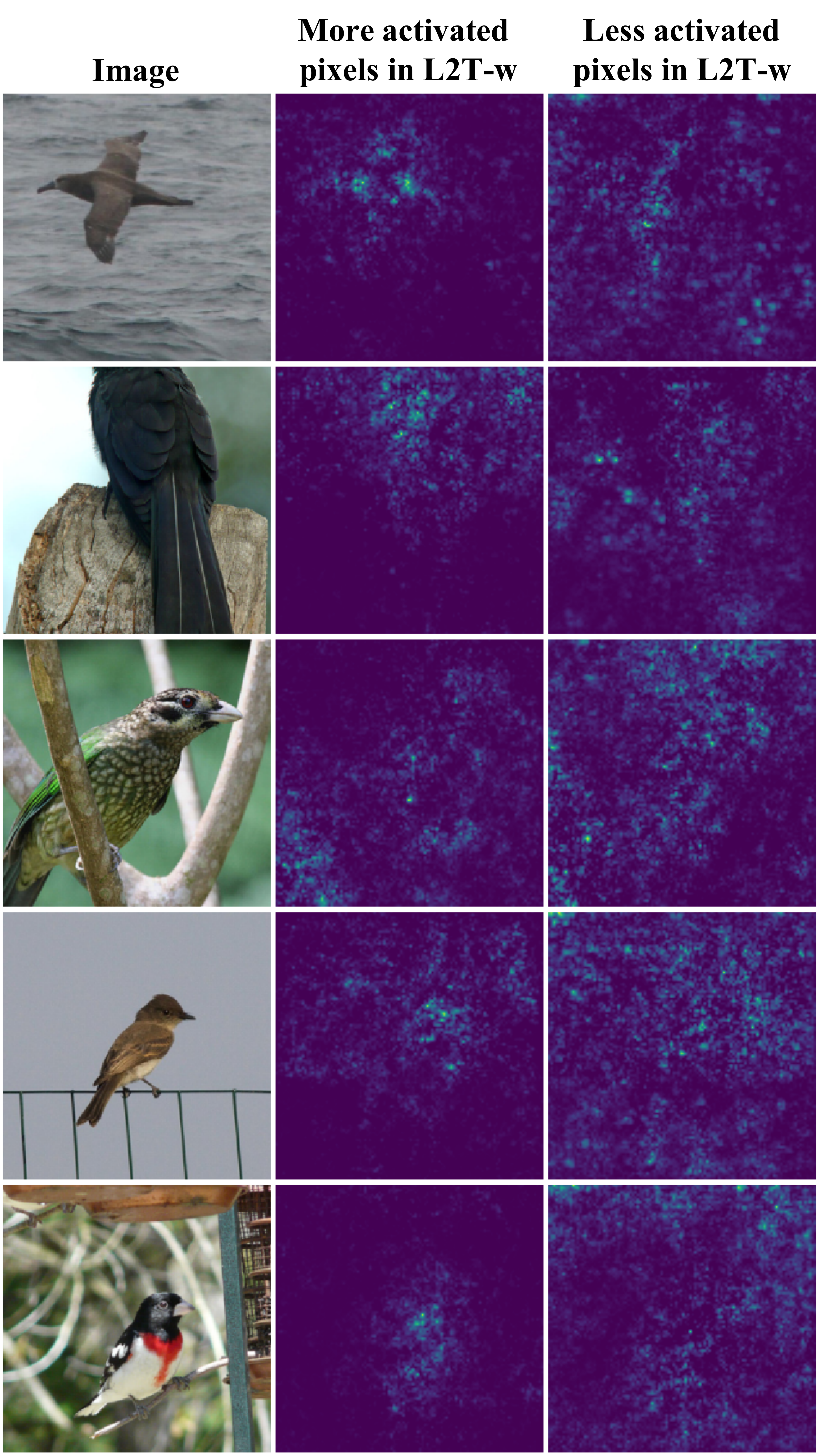}
\label{fig:cub_vis}}
\quad\quad
\subfigure[Stanford Dogs]{
\includegraphics[width=0.42\textwidth]{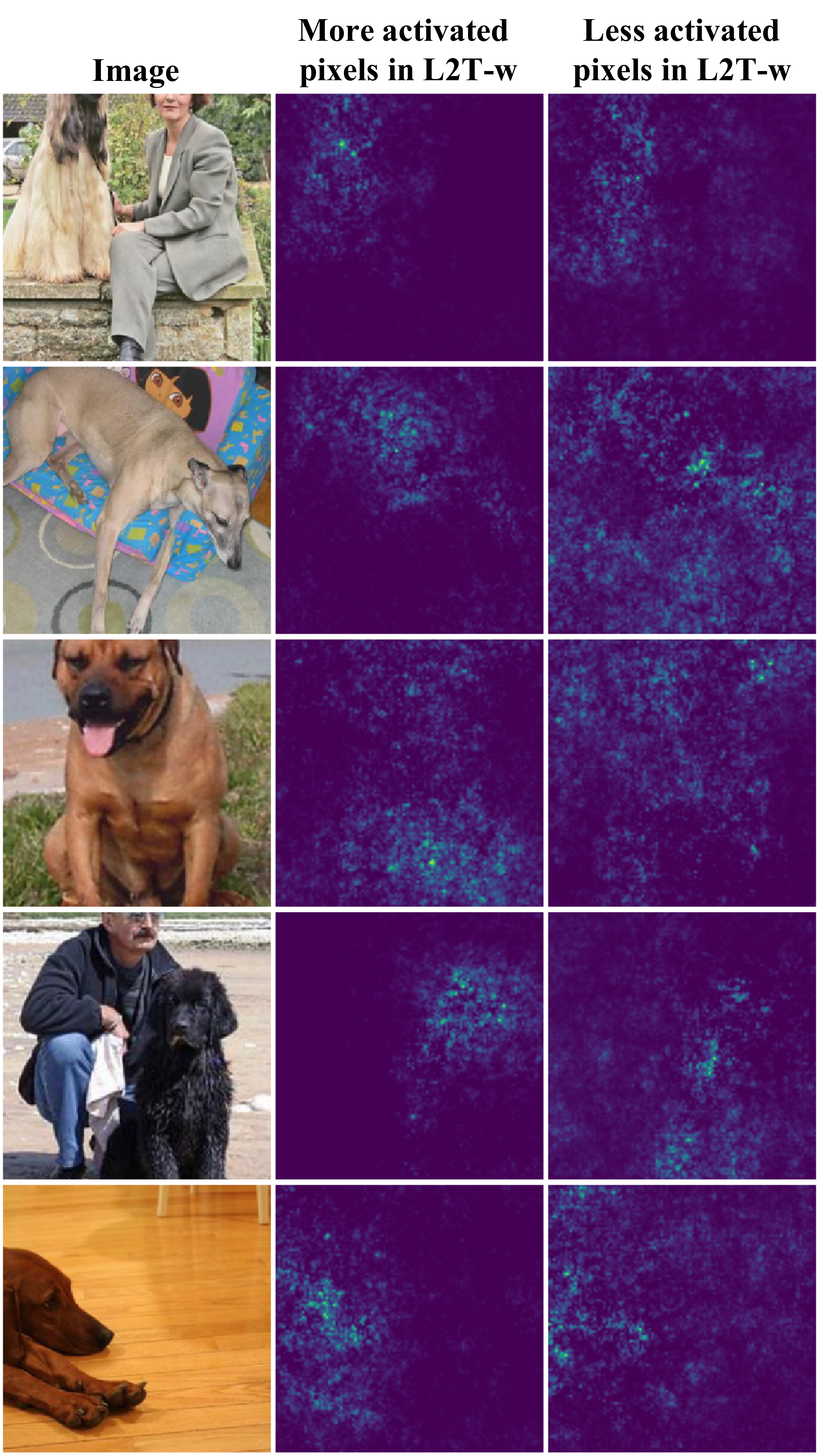}
\label{fig:dog_vis}}
\vspace{-0.05in}
\caption{{More (the second column) and less (the third column) activated pixels in the saliency maps of L2T-w compared to unweighted
feature matching (FM) on images of (a) CUB200 and (b) Stanford Dogs datasets. When computing saliency maps, we use normalized gradients.
One can observe that the higher activated pixels induced by L2T-w
tend to correspond to where task-specific objects are, while less activated location spread over entire location.
}}\label{fig:vis}
\vspace{-0.05in}

\end{figure*}
}
\subsection{Evaluation on Various Target Tasks}

We first evaluate the effect of learning to transfer what (L2T-w)
without learning to transfer where.
To this end, we use conventional hand-crafted matching configurations, single and one-to-one,
illustrated in Figure \ref{fig:resvgg-last} and \ref{fig:resvgg-parallel}, respectively.
For most cases reported in Table \ref{tbl:tiny}, L2T-w
improves the performance on target tasks
compared to the unweighted counterpart (FM):
for fine-grained target tasks transferred from ImageNet,
the gain of L2T-w over FM is more significant.
The results support that our method, learning what to transfer, is more effective when target tasks have specific types of input distributions, e.g., fine-grained classification, while the source model is trained on a general task.

Next, instead of using hand-crafted matching pairs of layers, we also learn where to transfer starting from all matching pairs illustrated in Figure \ref{fig:resvgg-all-before}.
The proposed final scheme in our paper, learning to transfer what and where (L2T-ww),
often improves the performance significantly
compared to the hand-crafted matching (L2T-w).
As a result, L2T-ww
achieves the best accuracy for all cases (with large margin) reported in Table \ref{tbl:tiny}, e.g., on the CUB200 dataset, we attain $10.4\%$ relative improvement compared to the second best baseline.

Figure \ref{fig:resvgg-all} shows the amounts $\lambda^{m,n}$ of transfer
between pairs of layers after learning
transfer from TinyImageNet to STL-10.
As shown in the figure, our method transfers knowledge to higher layers in the target model:
$\lambda^{2,5}=1.40,~\lambda^{1.5}=2.62,~\lambda^{3,4}=2.88,~\lambda^{2,4}=0.74$.
The amounts $\lambda^{m,n}$ of other pairs are smaller than $0.1$, except $\lambda^{1,2}=0.21$.
Clearly, those matching pairs are not trivial to find by hand-crafted tuning,
which justifies that our method for learning where to transfer is useful.
Furthermore, since our method outputs sample-wise $\lambda^{m,n}$, amounts of transfer are adjusted
more effectively compared to fixed matching pairs over all the samples. 
For example, amounts of transfer from source features $S^{1}(x)$ have relatively
smaller variance over the samples (Figure~\ref{fig:1st_lw}) compared to the
those of $S^{3}(x)$
(Figure~\ref{fig:3rd_lw}).
This is because
higher-level features are more task-specific while lower-level features are more task-agnostic. It evidences that meta-networks $g_\phi$ adjust the amounts of transfer for each sample
considering the relationship between tasks
and the levels of abstractions of features.

\subsection{Experiments on Limited-Data Regimes}
When a target task has a small number of labeled samples for training,
transfer learning can be even more effective.
To evaluate our method (L2T-ww) on such limited-data scenario,
we use CIFAR-10 as a target task dataset by reducing the number of samples.
We use $N\in\{50, 100, 250, 500, 1000\}$ training samples for each class,
and compare the performance of
learning from scratch, LwF, AT, LwF+AT and
L2T-ww.
The results are reported in Figure~\ref{fig:fewshot}.
They show that
ours achieves significant
more improvements compared to other baselines,
when the volume of the target dataset is smaller.
For example,
in the case of $N=50$, our method achieves $64.91\%$ classification accuracy,
while the baselines, LwF+AT, AT, LwF and scratch
show $53.76\%$, $51.76\%$, $43.32\%$ and $39.99\%$, respectively.
Observe that
ours needs only $50$ samples per class to achieve
similar accuracy of LwF with 250 samples per class.

\subsection{Experiments on Multi-Source Transfer}
In practice, one can have multiple pre-trained source models with various source datasets.
Transfer from multiple sources may potentially provide more knowledge for learning a target task,
however, using them simultaneously could require more hand-crafted configurations of transfer, such as balancing the transfer from many sources or choosing different pairs of layers depending on the source models.
To evaluate the effects of using multiple source models,
we consider the scenarios transferred from two source models simultaneously,
where the models are different architectures (ResNet20, ResNet32) or
trained on different datasets (TinyImageNet, CIFAR-10).
In Table \ref{tbl:multisource-v2},
we report the results of ours (L2T-ww) and other transfer methods
on a target task STL-10 with 9-layer VGG as a target model architecture.

Our method consistently improves the target model performance over
more informative transitions (from left to right in Table \ref{tbl:multisource-v2}) on sources, i.e.,
when using a larger source model (ResNet20 $\rightarrow$ ResNet32)
or using a different second source dataset (TinyImageNet $\rightarrow$ CIFAR-10).
This is not the case for all other methods.
In particular,
compare the best performance of each method transferred from
two TinyImageNet models and TinyImageNet+CIFAR-10 models as sources.
Then, one can conclude that
ours is the only one that effectively aggregates the heterogeneous
source knowledge, i.e., TinyImageNet+CIFAR-10.
It shows the importance of choosing the right configurations of transfer
when using multiple source models, and confirms that
ours can automatically decide the
useful configuration from many possible candidate pairs for transfer.

\subsection{Visualization}
With learning what to transfer, our weighted feature matching will allocate larger attention to task-related channels of feature maps. To visualize the attention used in knowledge transfer,
we compare saliency maps \citep{simonyan2013deep} for unweighted (FM) and weighted (L2T-w) matching
between the last layers of source and target models.
Saliency maps can be computed as follows:
\begin{align*}
    M_{i,j}=\max_c\left|\frac{\partial\mathcal{L}_{\tt wfm}^{m,n}(\theta|x,w^{m,n})}{\partial x_{c,i,j}}\right|
\end{align*}
where $x$ is an image, $c$ is a channel of the image, e.g., RGB, and
$(i,j)\in \{1,2,\dots, H\}\times\{1,2,\dots, W\}$
is a pixel position.
For the unweighted case, we use uniform weights.
On the other hand,
for the weighted case,
we use the outputs $w^{m,n}=f^{m,n}_\phi(S^{m}(x))$
of meta-networks learned by our meta-training scheme.
Figure \ref{fig:vis} shows which pixels are more or less activated in the saliency map
of L2T-w compared to FM. As shown in the figure,
pixels containing task-specific objects (birds or dogs) are more activated when using L2T-w, while background pixels
are less activated.
It means that the weights $w^{m,n}$ make knowledge of the source model be more
task-specific, consequently it can improve transfer learning.

%% file: Conclusion.tex
\section{Conclusion}\label{sec:con}
We propose a transfer method based on meta-learning which can transfer knowledge
selectively depending on tasks and architectures.
Our method transfers more important knowledge for learning a target task,
with
identifying \emph{what} and \emph{where} to transfer
using meta-networks.
To learn the meta-networks,
we design an efficient meta-learning scheme which requires
a few steps in the inner-loop procedure.
By doing so, we jointly train the target model and the meta-networks.
We believe that our work would shed a new angle
for complex transfer learning tasks between
heterogeneous or/and multiple network architectures and tasks.

%% file: supp.tex
\onecolumn
\clearpage
\begin{center}{\bf {\LARGE Supplementary Material:}}
\end{center}
\begin{center}{\bf {\Large Learning What and Where to Transfer}}
\end{center}

\appendix

\section{Network Architectures and Tasks}\label{sec:supp:arch_and_task}
For small image experiments ($32\times32$), we use TinyImageNet\footnote{https://tiny-imagenet.herokuapp.com/} as a source task,
and use
CIFAR-10, CIFAR-100~\citep{krizhevsky2009learning} and STL-10~\citep{coates2011analysis} datasets as target tasks.
CIFAR-10 and CIFAR-100 have 10 and 100 classes containing $5000$ and $500$ training images for each class, respectively, and each image has $32\times32$ pixels.
STL-10 consists of 10 classes, with 500 labeled images per each class in the training set.
Since the original images in TinyImageNet and STL-10 are not $32\times32$, we resize them into $32\times 32$ when training and testing.
We use a pre-trained 32-layer ResNet \cite{he2016deep} on TinyImageNet as a source model, and
we train 9-layer VGG~\citep{simonyan2014very}, which is the modified architecture used in Srinivas \& Fleuret~\yrcite{pmlr-v80-srinivas18a},
on CIFAR-10/100 and STL-10 datasets.

For large image experiments ($224\times224$),
we use a pre-trained 34-layer ResNet on ImageNet~\citep{deng2009imagenet} as a source model,
and consdier Caltech-UCSD Bird 200~\citep{WahCUB_200_2011}, MIT Indoor Scene Recognition~\citep{quattoni2009recognizing},
Stanford 40 Actions~\citep{yao2011human}
and Stanford Dogs~\citep{stanford_dogs} datasets as target tasks.
Caltech-UCSD Bird 200 (CUB200) contains 5k training images of 200 bird species.
MIT Indoor Scene Recognition (MIT67) has 67 labels for indoor scenes and 80 training images per each label. Stanford 40 Actions (Stanford40) contains 4k training images of 40 human actions. Stanford Dogs has 12k training images of 120 dog categories.
For these target fine-grained datasets, we train 18-layer ResNets.

\section{Optimization}\label{sec:supp:opt}
All target networks are trained by stochastic gradient descent (SGD) with a momentum of $0.9$.
We use a weight decay of $10^{-4}$
and an initial learning rate $0.1$ and decay the learning rate with a cosine annealing \cite{loshchilov2016sgdr}:
$\alpha_t=\frac{1}{2}(1+\cos\frac{t}{T}\pi)$ where $\alpha_t$ is the learning rate at epoch $t$, and $T$ is the maximum epoch.
For all experiments, we train target networks for $T=200$ epochs.
The size of mini-batch is $128$ for small image experiments, e.g., CIFAR,
or $64$ for large image experiments, e.g.,
CUB200.
When using feature matching, we use $\beta=0.5$.
For data pre-processing and augmentation schemes, we follow He et al.~\yrcite{he2016deep}.
We use the ADAM \citep{kingma2014adam} optimizer for training
the meta-networks $f_\phi$, $g_\phi$ with a learning rate of $10^{-3}$ or $10^{-4}$,
and a weight decay of $0$ or $10^{-4}$.
In our meta-training scheme, we observe that $T=2$
is enough to learn what and where to transfer.
We repeat experiments 3 times and report the average performance as well as the standard deviation.

\section{Ablation Studies}
\subsection{Comparison between the meta-networks and meta-weights}
\begin{table}[h]
\centering
\caption{Classification accuracy (\%) of transfer learning using meta-networks or meta-weights.}\label{tbl:meta-weights}
\vspace{0.1in}
\begin{tabular}{ccccc}
\toprule
Target task   & CUB200 & MIT67 & Stanford40 \\ \midrule
meta-weights  & 61.75 & 64.10 & 58.88 \\
meta-networks & 65.05 & 64.85 & 63.08 \\ \bottomrule
\end{tabular}
\end{table}

The weights, channel importance $w^{m,n}$ and connection importance
$\lambda^{m,n}$, decide amounts of transfer given a sample to
meta-networks. One can also learn directly $w^{m,n}$ and
$\lambda^{m,n}$ as constant meta-weights  using suggested bilevel scheme without
meta-networks.
Here, we compare the effectiveness of using \emph{meta-networks},
which gives different amount of transfer for each sample,
to learning \emph{meta-weights} directly, giving
the same importance over all the samples.
For fair comparison, we use same hyperparameters as described in Section \ref{sec:supp:arch_and_task} and \ref{sec:supp:opt},
except the meta-parameters.
As reported in Table \ref{tbl:meta-weights},
the performance of target models using meta-networks outperforms
the one using meta-weights up-to 4.2\%, which supports the effectiveness of using selective transfer
depending on samples.

\subsection{Comparison between the proposed bilevel scheme and original one}
To validate the effectiveness of the suggested bilevel scheme,
we perform experiments comparing the performance of target models trained with meta-networks,
using the proposed and original bilevel scheme.
For a fair comparison, we use $T=2$ for both methods,
and the other hyperparameters, model architectures and the source task are same with
the ones in Section \ref{sec:supp:arch_and_task} and \ref{sec:supp:opt}.
\begin{table}[h]
\centering
\caption{Classification accuracy (\%) of transfer learning using the original or proposed bilevel schemes.}\label{tbl:bilevel}
\vspace{0.1in}
\begin{tabular}{cccc}
\toprule
Target task & CUB200 & MIT67 & Stanford40 \\ \midrule
Original    & 35.38  & 54.18 & 53.47      \\
Ours        & 65.05  & 64.85 & 63.08      \\ \bottomrule
\end{tabular}
\end{table}
The original scheme obtains significantly lower accuracies
than the proposed bilevel scheme (Table \ref{tbl:bilevel}).
With much larger $T$, e.g., 5$\sim$100, a target model with the original bilevel scheme
does not succeed to obtain comparable performance with our bilevel scheme.
Moreover the meta-training time for meta-networks is increasing linearly as $T$ increases,
thus the original scheme is not applicable to practical scenarios.
These results show that the proposed bilevel scheme is more effective for learning meta-networks for selective transfer.